\documentclass[runningheads]{llncs}

\usepackage[T1]{fontenc}
\usepackage{multirow}
\usepackage{graphicx}
\begin{document}
\title{Multiple weather images restoration using the task transformer and adaptive mixup strategy}
%
%
\author{Yang Wen\inst{1}\orcidID{0000-0001-6303-8178} \and
Anyu Lai\inst{1} \and
Bo Qian\inst{2} \and Hao Wang\inst{1} \and Wuzhen Shi\inst{1} \and Wenming Cao\inst{1}}

\authorrunning{Yang Wen \and Anyu Lai et al.}
%
\institute{Guangdong Provincial Key Laboratory of Intelligent Information Processing, School of Electronic and Information Engineering, Shenzhen University, Shenzhen, China \and
 The Department of Comupter Science and Engineering, Shanghai Jiao Tong University, Shanghai, China}
\maketitle              
\begin{abstract}
The current state-of-the-art in severe weather removal predominantly focuses on single-task applications, such as rain removal, haze removal, and snow removal. However, real-world weather conditions often consist of a mixture of several weather types, and the degree of weather mixing in autonomous driving scenarios remains unknown. In the presence of complex and diverse weather conditions, a single weather removal model often encounters challenges in producing clear images from severe weather images. Therefore, there is a need for the development of multi-task severe weather removal models that can effectively handle mixed weather conditions and improve image quality in autonomous driving scenarios. In this paper, we introduce a novel multi-task severe weather removal model that can effectively handle complex weather conditions in an adaptive manner. Our model incorporates a weather task sequence generator, enabling the self-attention mechanism to selectively focus on features specific to different weather types. To tackle the challenge of repairing large areas of weather degradation, we introduce Fast Fourier Convolution (FFC) to increase the receptive field. Additionally, we propose an adaptive upsampling technique that effectively processes both the weather task information and underlying image features by selectively retaining relevant information. Our proposed model has achieved state-of-the-art performance on the publicly available dataset.

\end{abstract}
\section{Introduction}
The removal of adverse weather conditions, including rain, snow, and haze, from images is a critical challenge in numerous fields. Extreme weather events significantly impair the ability of computer vision algorithms to extract relevant information from images. Therefore, mitigating such weather-related effects is necessary to enhance the reliability of computer vision systems\cite{Du_Xu_Zhen_Cheng_Shao_2020}\cite{Ren_Zuo_Hu_Zhu_Meng_2019}\cite{Yang_Tan_Wang_Fang_Liu_2021}.

In this paper, we propose a novel approach for recovering multi-weather degraded images by leveraging a task sequence generator and an adaptive module. During the feature extraction phase, we employ a Task Intra-patch Block (TIPB) to partition the image into smaller patches and extract degraded features from them. These features are utilized not only in subsequent feature extraction stages but also fed into a Task Query Generator to generate task sequences based on the input task characteristics at each stage. This enables us to selectively focus on different types of degradation information during the upsampling stage. To address the challenge of handling large-area degraded features, we utilize Fast Fourier Convolution (FFC) to expand the receptive field. Finally, to fuse degraded information with background information, we employ adaptive upsampling techniques in the image restoration process. Our main contributions are:

\begin{itemize}
    \item We introduce a novel and highly efficient solution for tackling severe weather removal challenges, specifically focusing on image inpainting guided by the generation of weather degradation information and task feature sequences. Our proposed method surpasses the performance of existing state-of-the-art approaches in both real-world datasets and downstream object detection tasks.
    \item We propose the Task Intra-patch Block (TIPB), a novel feature extraction block that effectively captures detailed features of various degradation types at different scales. By utilizing TIPB at multiple stages, our approach can extract highly informative features that are tailored to each stage of the image restoration process. This enables us to effectively address different types of degradation and achieve superior performance in restoring degraded images.
    \item We present a novel task sequence generator that leverages multi-scale degradation details to generate task feature sequences. Our approach effectively captures the complex relationships between different degradation types at different scales and generates task sequences that are tailored to the specific. characteristics of the input image.
\end{itemize}

\section{Related work}

Deep learning-based solutions have become increasingly popular for various weather-related image restoration tasks, including rain removal\cite{YiyangShen2022SemiDRDNetSD}, hazy removal\cite{YeyingJin2022StructureRN}, and snow removal\cite{KaihaoZhang2021DeepDM}. These approaches have demonstrated significant performance improvements compared to traditional methods.

\subsection{For Rain Removal}
 Jiang et al.\cite{KuiJiang2020MultiScalePF} mainly proposed a multi-scale progressive fusion network (MSPFN). For the imaging principle of rain, due to the different distances between the rain and the camera, the rain in the image will show different ambiguities and resolutions, so the complementary information between multi-resolution and multi-pixels can be used to represent rain streaks The main paper proposes a framework from the perspective of input geometry and depth graphics, explores the multi-geometry representation of rain streaks, and first accomplishes deraining. For pixel rain streaks at different locations, the gradient calculation is used to obtain the global texture, so as to explore the complementarity in the spatial dimension and read out information to characterize the target rain streaks.

\subsection{For Haze Removal} To address the dense and uneven distribution of haze, Jin et al.\cite{YeyingJin2022StructureRN} propose a model that extracts feature representations from a pre-trained visual transformer (DINO-ViT) to recover background information. To guide the network to focus on non-uniform haze regions and then remove the haze accordingly, they introduce uncertainty feedback learning, which produces uncertainty maps with higher uncertainty in denser haze regions and can be viewed as the attention map representing the density of the haze and the non-uniformity of the distribution. Let the feedback network iteratively refine our dehazing output based on the uncertainty map. 

 \subsection{For Snow Removal} Currently, handcrafted features are still the mainstream for snow removal, making it difficult to achieve large-scale generalization. Liu et al.\cite{YunFuLiu2017DesnowNetCD} designed a multi-stage network called DesnowNet to sequentially handle the removal of translucent and opaque snow particles. We also differentiate snow translucency and color difference properties for accurate estimation. Furthermore, their method separately estimates the remaining complement of snow-free images to recover details occluded by opaque snow. In addition, the whole network adopts a multi-scale design to simulate the diversity of snow.

\subsection{Multi-task Weather-related Image Restoration}
Li et al.\cite{RuotengLi2020AllIO} first designed a generator with multiple task-specific encoders, each associated with a specific type of severe weather degradation. They utilize neural architecture search to optimally process image features extracted from all encoders. Subsequently, to transform degraded image features into clean background features, they introduce a series of tensor-based operations that encapsulate the fundamental physics behind the formation of rain, haze, snow, and adherent raindrops. These operations are the basic building blocks of schema search. Finally, the discriminator simultaneously evaluates the correctness of the restored image and classifies the degradation type. Valanarasu et al.\cite{JeyaMariaJoseValanarasu2022TransWeatherTR} propose TransWeather, a Transformer-based end-to-end model that can recover images degraded by any weather condition with only one encoder and one decoder. Specifically, they exploit a novel Transformer encoder that uses intra-patch Transformer blocks to enhance intra-patch attention to effectively remove small weather degradations.

\section{Method}

\begin{figure*}
\begin{center}  
    \includegraphics [width=1\textwidth]{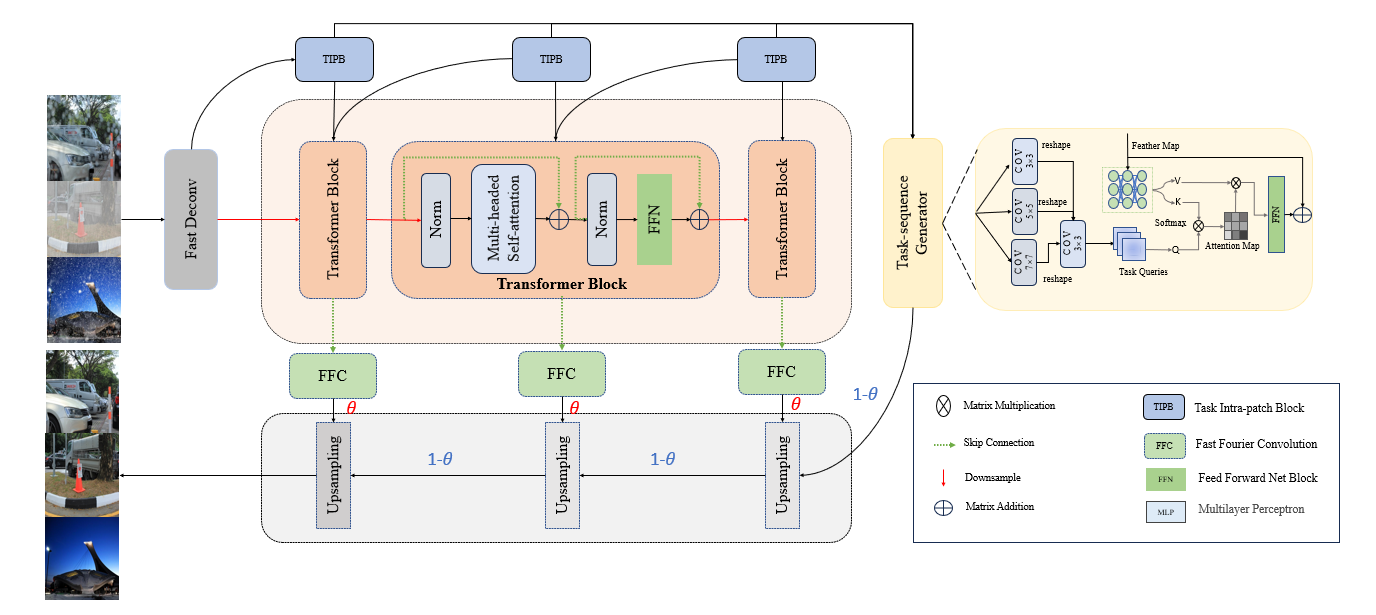}
    \end{center}
\caption{Overview of the proposed network. The degraded image will first be input to the deconvolution module for preliminary processing, and then the task features will be extracted layer by layer through the TIPB module and the transformer module, and the task features will be input to the Task Sequence Generator to generate a task sequence. Finally, the background features from the bottom layer will be combined with the task features through the FFC module to restore a clear image. }
\label{fig_2}
\end{figure*}

We propose a novel framework to tackle different image degradation tasks, as shown in Figure\ref{fig_2}. In this section, we provide a comprehensive overview of the network framework.

\subsection{Network Architecture}
The proposed network takes a 3×3 weather-degraded image as input. The input image will be processed by a multi-stage Transformer block to generate local information from different stages. The output of each stage is then input to the TIPB module, which extracts degradation-specific detail features according to the specific degradation task. The resulting multi-stage degradation features are fed into the task sequence generator to generate a task sequence that facilitates the identification of the specific degradation task affecting the input image. To capture more global information, the FFC module is introduced into the network. The down-sampled output of each stage is first input to the FFC module to extract global information, which is then used to assist in image restoration in the subsequent upsampling stage. In the upsampling stage, learnable parameters are employed to selectively fuse task features and image features, ultimately resulting in the restoration of a clear image.

\subsection{Task Intra-patch Block}
\begin{figure}[]
    \begin{center}
    \includegraphics [width=1\columnwidth]{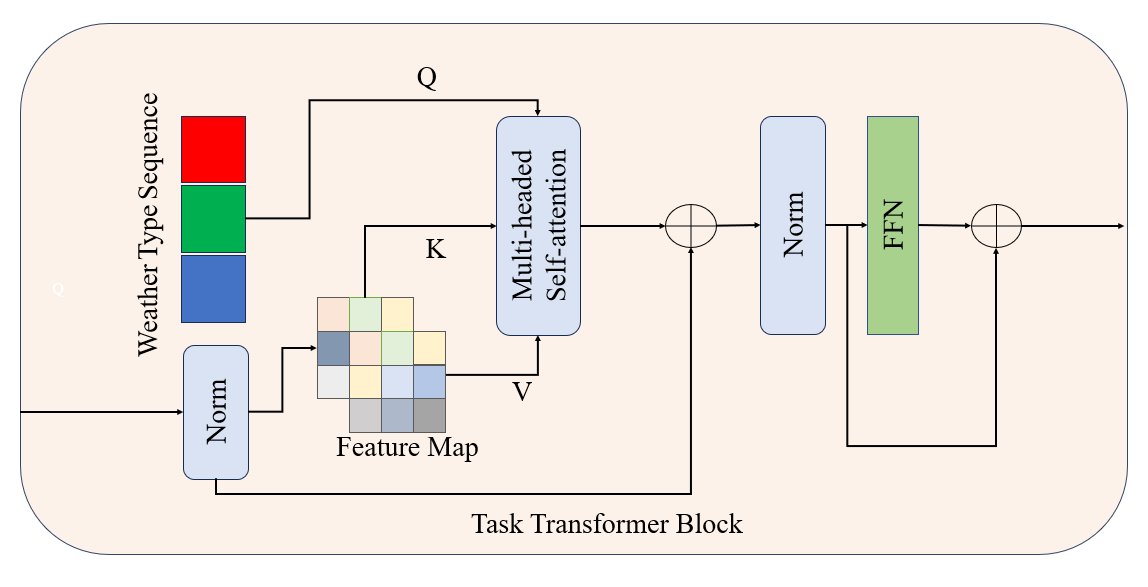}
    \end{center}
    \caption{Detail of Task Transformer Block. By calculating the q introduced from the outside and the kv generated by the image, the attention map is input to the multi-layer perceptron to obtain the feature map with task information.}
    \label{figure3}
    \end{figure}

At each stage, the Task Intra-patch Block(TIPB) processes the image features, which are first cropped to half the size of the original image to facilitate the extraction of smaller degraded details. As shown in Figure\ref{figure3}, in order to adaptively query different degraded features, an external learnable sequence is introduced and optimized during the training of the network. The resulting sequence generates a feature map that contains a substantial amount of task-specific information, which is combined with the input image and input into the Transform Block of the next stage. The feature maps from each stage are jointly input into the task sequence generator to generate a task query vector that assists in identifying the specific degradation task affecting the input image. This approach facilitates the effective extraction of task-specific information at each stage, ultimately leading to improved performance in image restoration. The output of TIPB can be expressed as:
\begin{equation}
    TIPB_{i}(I_i)=FFN(MSA(I_i)+I_i)
\end{equation}
where $T(\cdot)$ represents the transformer block, $FFN(\cdot)$ repre-
sents the feed-forward network block, $MSA(\cdot)$ represents
multi-head self-attention, $I$ is the input and $i$ represents the
stage in the encoder. The multi-head attention of the TIPB module is different from the traditional form, and its self-attention is defined as follows: 
\begin{equation}
    Attn(Q,K,V) = softmax(\frac{Q_{learnabled}K^T}{\sqrt{d}})V
\end{equation}
The proposed network leverages a randomly generated task query sequence ($Q$) to represent a diverse range of weather conditions. The keys ($K$) and values ($V$) used in the attention mechanism are derived from the input feature map.
\subsection{Task Sequence Generator}
The TIPB module introduces a stochastic task vector into the transformer module as a query in the attention mechanism. This vector is trained concurrently with the network and facilitates the capture of degradation characteristics under varying weather conditions. TIPB operates on each level of the encoder to extract degradation information of diverse scales in the image. The TIPB's output is subsequently fed into the encoder of the subsequent stage, and the outputs of all stages are jointly utilized as input to the Task Sequence Generator for generating a task sequence pertaining to the image.

The Task Sequence Generator comprises several convolutional layers of varying scales and a self-attention module. These convolutions, operating at different stages, enable effective processing of the output of the Task Information Processing Block (TIPB). We then utilize a 3x3 convolutional layer to combine the task information from the four different scales, resulting in a task feature query vector map. This map is subsequently used as the query (Q) in conjunction with the image in the self-attention mechanism to generate a feature map that contains rich task information. The output of Task sequence generator can be expressed as:
\textbf{\begin{equation}
    Tsg(I, Q_{Task})=FFN(MSA(I, Q_{Task})+I)
\end{equation}}

\textbf{\begin{equation}
   \small Q_{Task}=Cov_{3,3}(Cov_{7,7}(T_1)+Cov_{5,5}(T_2)+Cov_{3,3}(T_3))
\end{equation}}
Where $Tsg(\cdot)$ represents the output of the Task-sequence Generator, $FFN(\cdot)$, and $MSA(\cdot)$ represent the feedforward network and the multi-head self-attention module, respectively. I represent the feature map input to the Task-sequence Generator, and $Q_{Task}$ represents the generated task query sequence. $T_i$ denotes the output of TIPB from the i-th stage. $Conv_{n,n}$ represents the use of n×n convolution operations.

\begin{figure*}
\begin{center}
    \includegraphics [width=1\textwidth]{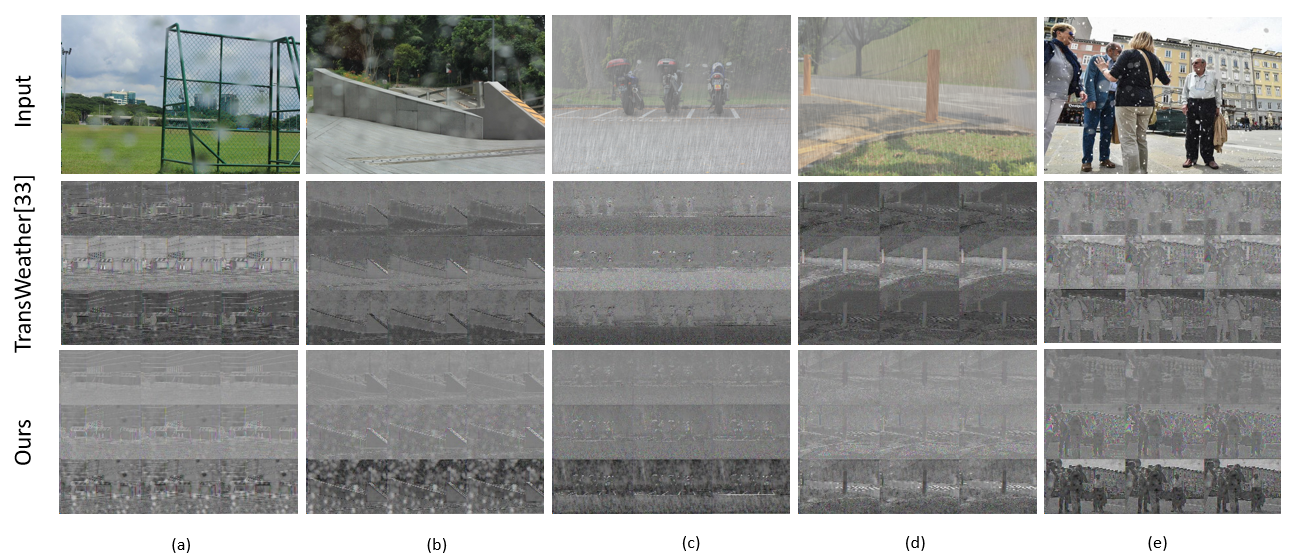}
    \end{center}
\caption {A comparison of the intermediate results of three tasks using the TransWeather\cite{JeyaMariaJoseValanarasu2022TransWeatherTR} Model and the Task-sequence Generator module. (a) and (b) is the output result of removing rain, (c) and (d) is the output result of removing rain and haze, and (e) is the output result of removing snow. }
\label{fig_3}
\end{figure*}

The Task-sequence Generator module improves the ability to capture the degradation characteristics of different weather conditions. Figure \ref{fig_3} presents a comparison of the intermediate results of three tasks using the Base Model and the Task-sequence Generator module. Specifically, outputs a and b correspond to the rain removal task, c and d correspond to the haze removal and rain removal task, and e corresponds to the snow removal task. The first three rows illustrate the input image, the output of the Base Model, and the output after incorporating the Task-sequence Generator module.

Our results demonstrate that the Task-sequence Generator module enhances the ability to capture degradation characteristics compared to the Base Model. Specifically, the degradation information in the output images is clearer, and the contrast between the degradation content and the background is stronger. These findings highlight the importance of incorporating advanced techniques, such as the Task-sequence Generator module, in image restoration tasks to improve performance and enhance the quality of the results.
\subsection{Fast Fourier Convolution }
 The restoration of weather images with dense degradation has posed a significant challenge in the field of image restoration. Conventional methods have primarily relied on local background information to restore detailed features, but their efficacy in dealing with large-scale degradation has been limited. Recently, Roman Suvorov et al. proposed a novel technique that leverages global information to tackle this problem. Building upon their work, we have employed this approach in the context of weather image restoration, enabling us to incorporate a wider range of background information and achieve effective restoration of images with large-scale degradation.

\begin{figure}[]
    \begin{center}
    \includegraphics [width=1\textwidth]{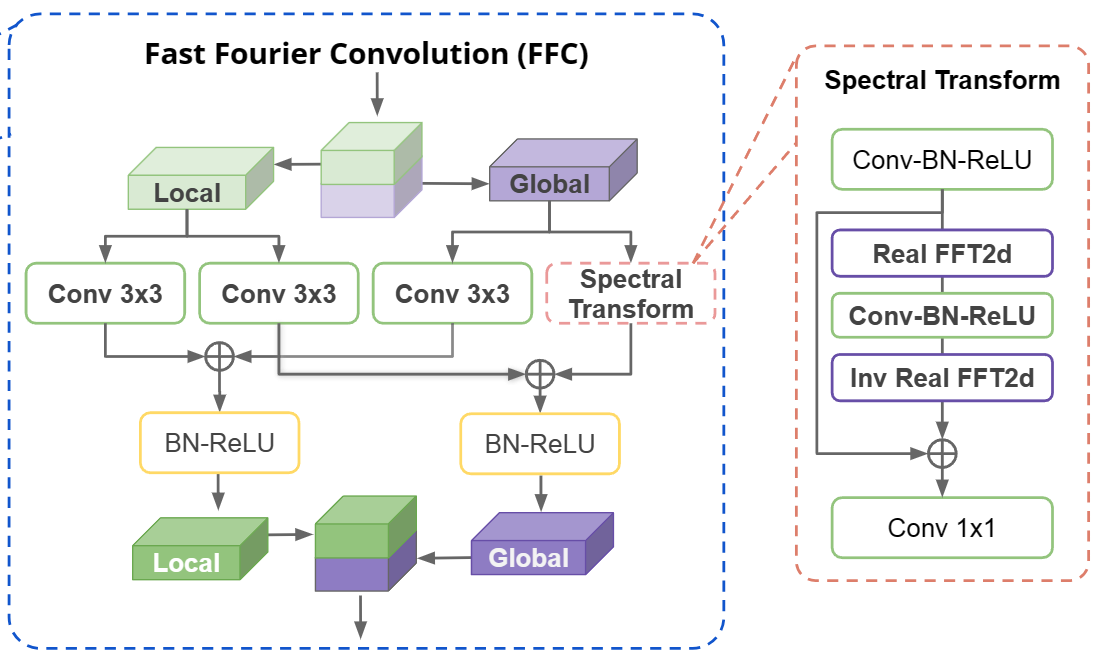}
    \end{center}
    \caption{Detail of Task FFC Block. The local branch utilizes conventional convolution operations, while the global branch employs channel-wise fast Fourier transform to capture the broader context of the image.}
    \label{figure5}
    \end{figure}
    
The illustration of FFC is available in Figure \ref{figure5}. The input feature map is split into two branches for parallel processing. The local branch performs a conventional convolution operation, while the global branch employs channel-wise fast Fourier transform (FFT) to capture the global context. The information from these two branches is then fused to generate a feature map with a receptive field that covers the entire image. This feature map is subsequently applied to the upsampling process, resulting in the restoration of a more realistic and detailed image. The steps of FFC are defined as follows:

a) applies Real FFT2d to an input tensor
  \begin{eqnarray}
    Real FFT2d: R^{H \times  W \times  C} \rightarrow C^{H \times  \frac{W}{2} \times  C}
\end{eqnarray}
and concatenates real and imaginary parts
\begin{eqnarray}
    Complex To Real: C^{H \times  \frac{W}{2} \times  C} \rightarrow R^{H \times  \frac{W}{2} \times  2C}
\end{eqnarray}

b) applies a convolution block in the frequency domain
 \begin{eqnarray}
    Rule \circ BN \circ Conv 1\times 1 : R^{H \times  \frac{W}{2} \times  2C} \rightarrow R^{H \times  \frac{W}{2} \times  2C}
\end{eqnarray}

c) applies inverse transform to recover a spatial structure
 \begin{eqnarray}
    Real To Complex : R^{H \times  \frac{W}{2} \times  2C} \rightarrow C^{H \times  \frac{W}{2} \times  C}
\end{eqnarray} 
    
 \begin{eqnarray}
    Inverse Real FFT2d  : C^{H \times  \frac{W}{2} \times  C} \rightarrow R^{H \times  W \times  C}
\end{eqnarray} 

Firstly, we apply the fast Fourier transform (FFT) to the input feature map. Next, we combine the real part and the imaginary part obtained from the FFT. Subsequently, we perform the convolution operation on the combined feature map, emphasizing the global background information. Finally, we separate the combined feature map into real and imaginary parts and restore the feature map to the time domain using the inverse Fourier transform.

\subsection{Adaptive Mixup For Feature Preserving}
The proposed network incorporates an encoder-decoder architecture that can effectively extract low-level features from the input image and task-specific features from the degraded image. Adaptive upsampling is utilized to enable the effective mixing of task information and image features. Addition-based skip connections, which are commonly used in encoder-decoder models, may lead to loss of shallow features or external task information. To address this, the Adaptive Mixup approach is introduced, which is able to retain more texture information of the image by adaptively mixing the features from different levels of the network. The output of Adaptive Mxiup can be expressed as:
   \small{ \begin{eqnarray}
        f_{\uparrow i+1} = Mix(f_{\downarrow m-i}, f_{\uparrow i})=\sigma (\theta _{i})*f_{\downarrow 
i}+(1-\sigma (\theta_{i}) *f_{\uparrow i})
    \end{eqnarray}}
Where $f_{\uparrow i}$ and $f_{\downarrow m-i}$ represent the upsampling and downsampling feature maps of the i-th stage($i\subseteq \{1,2...m$\}), $\sigma(\theta_{i})$ represents the learnable factor of the i-th stage, which is used to fuse the low-level features from the downsampling and the task features from the decoder, and Its value is determined by the sigmoid operator on the parameter $\theta_{i}$.

\section{Result}
In this section, we conducted an extensive experimental analysis to validate the effectiveness of our proposed approach. Specifically, we provide detailed information regarding the dataset used, experimental design, and comparative analysis with state-of-the-art techniques.

\subsection{Comparison with state-of-the-art}
We compare our method against state-of-the-art modalities specifically designed for each task. We compare with state-of-the-art methods such as Attention Gan\cite{RuiQian2017AttentiveGA} and Swin-IR\cite{JingyunLiang2021SwinIRIR}. 
At the same time, we also compare with the state-of-the-art methods of multi-tasking such as All-in-one\cite{RuotengLi2020AllIO} and Transweather\cite{JeyaMariaJoseValanarasu2021TransWeatherTR}. The detailed experimental results will be shown below.

 \begin{figure*}
\begin{center}
    \includegraphics [width=1\textwidth]{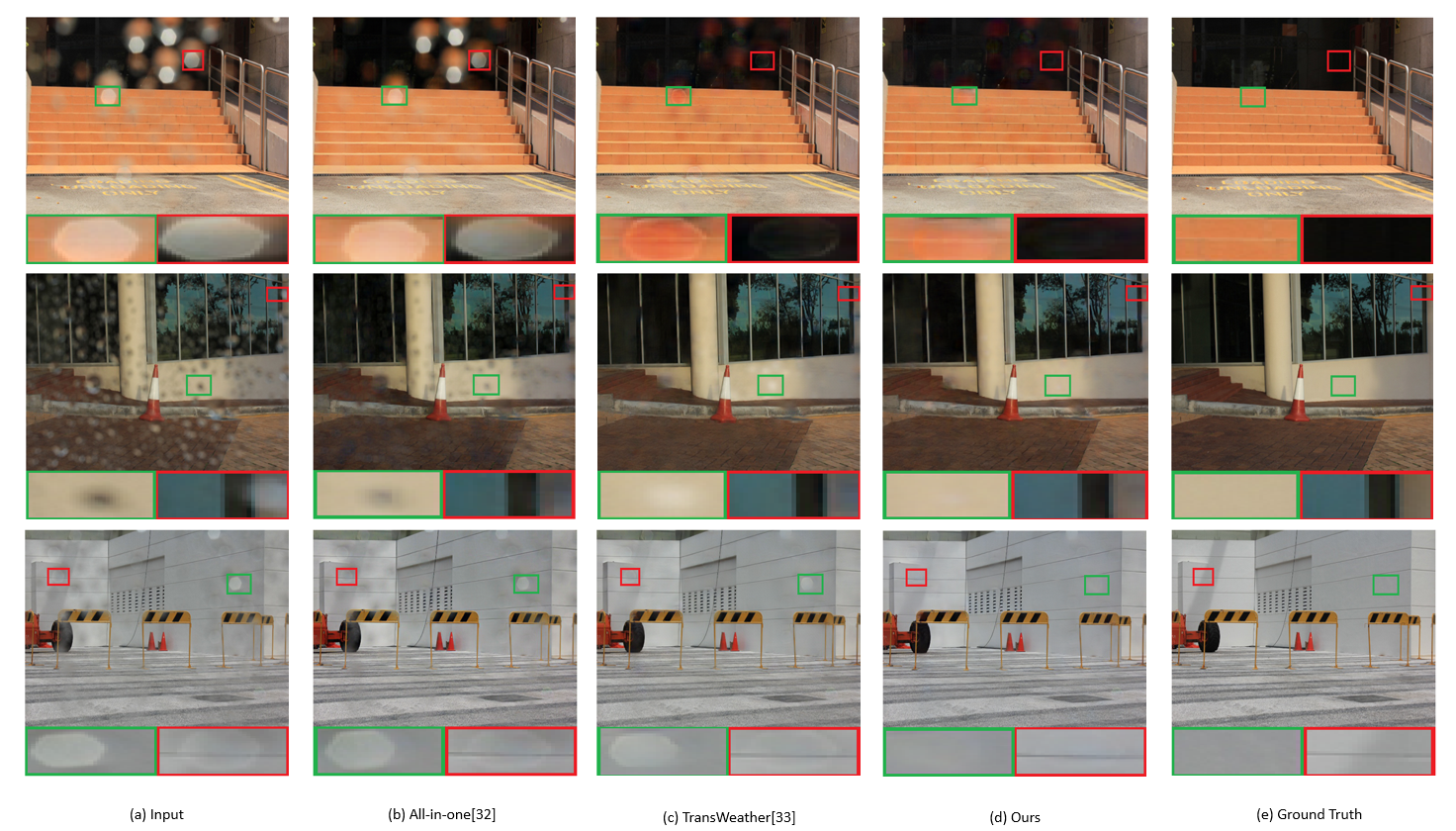}
    \end{center}
\caption {Qualitative results comparison of the proposed method with existing state-of-the-art methods All-in-One\cite{RuiQian2017AttentiveGA} and TransWeather\cite{JeyaMariaJoseValanarasu2022TransWeatherTR}, on Raindrop dataset\cite{RuiQian2017AttentiveGA} for raindrop removal.}
\label{figure6}
\end{figure*}
\subsubsection{Visual Quality Comparison}
We performed a qualitative comparison with All-in-One Network and TransWeather. The results are shown in Figures \ref{figure6}. Our proposed method exhibits superior performance in removing degradations, particularly in cases where there are large areas of degenerate features. In the first scenario of Figure \ref{figure6}, our ability to remove large raindrops is superior to the other two models. Furthermore, our approach outperforms other models in terms of restoring texture and color information in the image.  In the second scenario of the Figure\ref{figure6}, our model significantly outperforms the other two models in restoring the wall color. Both All-in one and TransWeather leave residual color changes caused by raindrops on the image.

\subsubsection{Referenced Quality Metrics}
We use PSNR and SSIM to quantitatively evaluate the performance of different models in RainDrop test sets. The experimental results are shown in Tables \ref{table3}. Our results demonstrate that our proposed method outperforms the multi-task single-job approach in the combination of three distinct weather types. Moreover, when compared to other multi-task models, our approach exhibits superior performance.

\begin{table*}[t]
\renewcommand\arraystretch{0.7}
\caption{Quantitative comparison on the RainDrop test dataset based on PSNR and SSIM. ↑ means the higher the better.
 }
 \centering
\begin{tabular}{p{2cm}|p{3cm}p{3cm}p{2cm}p{2cm}}
\hline
\hline
Type                                                                      & Method                 & Venue    & PSNR$\uparrow$            & SSIM$\uparrow$             \\ \hline
\multirow{5}{*}{\begin{tabular}[c]{@{}c@{}}Type  Specific\end{tabular}} &  pix2pix\cite{PhillipIsola2016ImagetoImageTW}                          & CVPR2017 & 28.02                                 & 0.8547                                 \\
                                                                           & Attn.GAN\cite{RuiQian2017AttentiveGA} & CVPR2018 & 30.55                                 & 0.9023                                 \\
                                                                           & Quan et al\cite{YuhuiQuan2019DeepLF}                       & ICCV2019 & 31.44                                 & 0.9263                                 \\
                                                                           & Swin-IR\cite{JingyunLiang2021SwinIRIR}                          & CVPR2021 & 30.82                                 & 0.9035                                \\ \hline
\multirow{5}{*}{\begin{tabular}[c]{@{}c@{}}Multi-Task\end{tabular}}                                               & All-in-One\cite{RuotengLi2020AllIO}             & CVPR2020 & 31.12          & 0.9268          \\
 & TransWeather\cite{JeyaMariaJoseValanarasu2022TransWeatherTR}           & CVPR2022 & 28.84          & 0.9527          \\& Zhen's\cite{Tan_Wu_Liu_Chu_Lu_Ye_Yu_2023} &TIP2023& 31.03&0.9228\\
                                                                          & AIRFormer\cite{Gao_Wen_Zhang_Zhang_Chen_Liu_Luo_2024} &TCSVT2023& \textbf{32.09} &0.9450
                                                                          \\ 
  & Ours                   & --       & 29.35 & \textbf{0.9574} \\ \hline
\hline
\end{tabular}
\label{table3}
\end{table*}

\subsubsection{Object Detection Comparison}
Severe weather conditions significantly impact the field of autonomous driving, particularly in the context of object detection from acquired images. The ability to accurately detect objects in these images is crucial for analyzing the current driving situation and making appropriate judgments. In this chapter, we utilize the YOLOV5 algorithm to perform object detection on images repaired by each of the considered models.
Table\ref{table5} presents the quantitative analysis outcomes of our object detection methodology on a dataset comprising 200 objects. In terms of detection accuracy, our approach surpasses the performance of the All-in-One\cite{RuotengLi2020AllIO} method. Furthermore, when compared to the TransWeather\cite{JeyaMariaJoseValanarasu2022TransWeatherTR} technique, our detected objects exhibit a higher level of confidence.

\begin{table*}[]
\centering
\caption{Quantitative comparison of object detection with TransWeather and All-in-one. Errors and Omissions are the sum of the number of false detections and missed detections. Average Confidence is the average confidence of the recognized objects.}
\resizebox{\textwidth}{!}{
\begin{tabular}{c|c|c}
\hline
\hline
             & Errors and Omissions & Average Confidence \\ \hline
Input        & 144                  & 0.184              \\
All-In-One\cite{RuotengLi2020AllIO}   & 18                   & 0.624              \\
TransWeather\cite{JeyaMariaJoseValanarasu2022TransWeatherTR} & 3                    & 0.648              \\
Ours         & 3                    & 0.659              \\
GroundTruth  & 0                    & 0.681\\
\hline
\hline
\end{tabular}
}
\label{table5}
\end{table*}

\section{Conclusion}
In this paper, we present a novel model for addressing the challenges posed by multi-weather degraded images. Our proposed approach involves leveraging trainable sequences to extract multi-scale features, which are subsequently utilized to generate task sequences specific to degradation-related tasks. These task sequences serve as guidance for the network, enabling selective focus on degradation information from different tasks. To capture broader background information and facilitate the restoration of large degraded regions, we introduce a Fast Fourier Convolution (FFC) module. This module effectively captures global contextual information, aiding in the recovery process. Additionally, we employ adaptive mixing to fuse features obtained from different modules, enhancing the overall performance of our method. Our proposed approach overcomes the limitation of single-task-specific networks, which often struggle to be practically deployed. When compared to other multi-task processing networks, our method exhibits superior capabilities in extracting information on various weather degradations while effectively repairing extensive degraded content. To validate the effectiveness of our proposed method, we conducted extensive evaluations on diverse datasets. The experimental results demonstrate the superior performance of our approach, surpassing many state-of-the-art methods in the field.

%
%
%
\bibliographystyle{splncs04}
\bibliography{CGI2024_192}

\end{document}